\documentclass[sigconf]{acmart}
\AtBeginDocument{%
  \providecommand\BibTeX{{%
    \normalfont B\kern-0.5em{\scshape i\kern-0.25em b}\kern-0.8em\TeX}}}

\copyrightyear{2023}
\acmYear{2023}
\setcopyright{acmlicensed}
\acmConference[SSDBM 2023]{35th International Conference on Scientific and Statistical Database Management}{July 10--12, 2023}{Los Angeles, CA, USA}
\acmBooktitle{35th International Conference on Scientific and Statistical Database Management (SSDBM 2023), July 10--12, 2023, Los Angeles, CA, USA}
\acmPrice{15.00}
\acmDOI{10.1145/3603719.3603739}
\acmISBN{979-8-4007-0746-9/23/07}

\usepackage{url}
         
\usepackage{amsmath,amssymb,amsfonts}
\usepackage{algorithmic}
\usepackage{graphicx}
\usepackage{textcomp}
\usepackage{float}
\usepackage{subfigure}
\usepackage{multirow}
\usepackage[ruled,linesnumbered]{algorithm2e}

\begin{document}
\title{Multi-representations Space Separation based Graph-level Anomaly-aware Detection}
\author{Fu Lin}
\affiliation{
  \institution{School of Computer Science\\Wuhan University}
  \city{Wuhan}
  \state{Hubei}
  \country{China}}
\email{linfu@whu.edu.cn}

\author{Haonan Gong}
\affiliation{
  \institution{School of Computer Science\\Wuhan University}
  \city{Wuhan}
  \state{Hubei}
  \country{China}}
\email{gonghaonan@whu.edu.cn}

\author{Mingkang Li}
\affiliation{%
  \institution{School of Computer Science\\Wuhan University}
  \city{Wuhan}
  \state{Hubei}
  \country{China}}
\email{mingkangLi@whu.edu.cn}

\author{Zitong Wang}
\affiliation{%
  \institution{School of Computer Science\\Wuhan University}
  \city{Wuhan}
  \state{Hubei}
  \country{China}}
\email{zitongwang@whu.edu.cn}

\author{Yue Zhang}
\affiliation{%
  \institution{School of Computer Science\\Wuhan University}
  \city{Wuhan}
  \state{Hubei}
  \country{China}}
\email{yuezhang@whu.edu.cn}

\author{Xuexiong Luo}
\affiliation{%
  \institution{School of Computing\\Macquarie University}
  \city{Sydney}
  \state{NSW}
  \country{Australia}}
\email{xuexiong.luo@hdr.mq.edu.au}

\renewcommand{\shortauthors}{Lin and Gong, et al.}

\begin{abstract}
Graph structure patterns are widely used to model different area data recently. How to detect anomalous graph information on these graph data has become a popular research problem. The objective of this research is centered on the particular issue that how to detect abnormal graphs within a graph set. The previous works have observed that abnormal graphs mainly show node-level and graph-level anomalies, but these methods equally treat two anomaly forms above in the evaluation of abnormal graphs, which is contrary to the fact that different types of abnormal graph data have different degrees in terms of node-level and graph-level anomalies. Furthermore, abnormal graphs that have subtle differences from normal graphs are easily escaped detection by the existing methods. Thus, we propose a multi-representations space separation based graph-level anomaly-aware detection framework in this paper. To consider the different importance of node-level and graph-level anomalies, we design an anomaly-aware module to learn the specific weight between them in the abnormal graph evaluation process. In addition, we learn strictly separate normal and abnormal graph representation spaces by four types of weighted graph representations against each other including anchor normal graphs, anchor abnormal graphs, training normal graphs, and training abnormal graphs. Based on the distance error between the graph representations of the test graph and both normal and abnormal graph representation spaces, we can accurately determine whether the test graph is anomalous. Our approach has been extensively evaluated against baseline methods using ten public graph datasets, and the results demonstrate its effectiveness. The code for our method is publicly available on \url{https://github.com/whb605/MssGAD.git}
\end{abstract}



\maketitle

\section{Introduction}
The application of graph structure pattern \cite{zhang2018network,luo2020deep} is widely exploited in various scenarios. For example, people are associated with performing various social activities on social networks, and every user is affiliated with rich profile information. So users are as attributed nodes and relationships of users are as edges when social networks are treated as graphs. As the population of various graphs, graph anomaly detection application \cite{ma2021comprehensive,akoglu2015graph} gets increasing attention in recent years. Graph anomaly detection primarily aims to pinpoint uncommon nodes or graphs that exhibit significant deviations from the majority of other nodes or graphs within a given set of graph data. Thus, in this work, we concentrate on detecting abnormal graphs within a graph set, i.e., graph-level anomaly detection(GLAD), which is also especially significant in real-life applications. As an illustration, GLAD can be utilized to identify unusual molecules in extensive collections of molecules that are depicted as molecular graphs, where atoms are illustrated as nodes and bonds are represented as edges. This is because abnormal molecules have distinct structures and features in their corresponding graphs compared to other molecules. Similarly, compared with traditional financial fraud detection technology \cite{dou2020enhancing,zhang2022efraudcom}, GLAD can not only intuitively present the complex topology implied in the data, but also integrate the correlation between data objects into the fraud identification task, making it easier to identify the fraud that is hard to find. In conclusion, GLAD is more worth developing to improve the performance of anomalous graph detection. 

\begin{figure}[!h]
	\small
	\includegraphics[width=\linewidth]{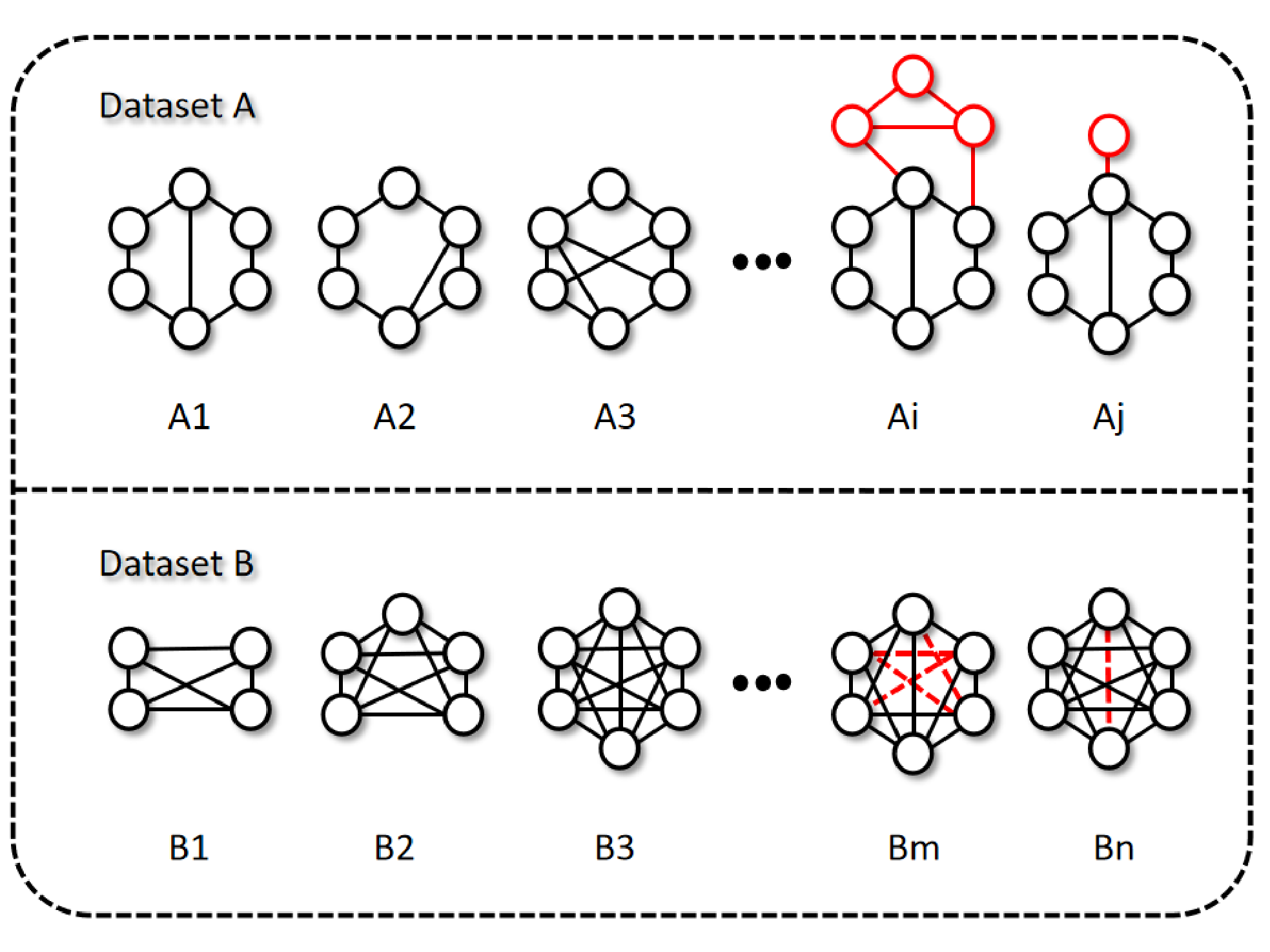}
	\caption{The example of different anomaly patterns in datasets. In dataset A, the overall structure of a normal graph is a hexagon connected by six nodes, and the edges between the nodes can be arbitrary. However, the difference in the anomaly graph is that this hexagonal structure is broken by the extra connected nodes, which indicates that the graph anomaly in dataset A needs to be judged in conjunction with the overall structure, which is also known as the graph-level anomaly. In contrast, the overall shape of the nodes connected in the normal graph in dataset B is not fixed, but there needs to be a connection linking each pair of nodes. While in the abnormal graph, there are no connected edges between several nodes. Therefore, the anomaly of dataset B is mainly manifested as the different connectivity of the nodes, that is, the node-level anomaly. In addition, even if they are both graph-level anomalies or node-level anomalies, the degree of significance of the difference between them and normal graphs is not the same. In dataset A, the structure of graph Aj is closer to a normal graph than graph Ai, while in dataset B, graph Bn also has fewer missing edges than graph Bm. These anomalous individuals that are more like the normal graphs will be more difficult to discriminate.}\label{fig1}
\end{figure}

In recent years, GLAD has gradually received attention and some methods have achieved good performance. Several studies have shown that GLAD can achieve promising results by utilizing off-the-shelf methods like one-class support vector machine (OCSVM)\cite{scholkopf1999support} and isolation forest (iForest)\cite{liu2008isolation}. However, the efficacy of these detection methods is restricted, because these measures are not good at extracting effective node-level and graph-level representations on complex graph data and the evaluation strategy is shallow. Due to the remarkable development of deep learning, some commonly used graph deep learning methods \cite{wu2020comprehensive} are migrated to GLAD and have shown excellent performance. such as \cite{chen2017outlier, hawkins2002outlier, pang2018learning, pang2019deep, perera2019learning}. These approaches initially investigate the observation that abnormal graphs consistently manifest anomalies at both the node and graph levels. Subsequently, these methods employ encoders based on graph neural networks (GNNs) \cite{kipf2016semisupervised,velivckovic2017graph,hamilton2017inductive} to learn node-level and graph-level representations of normal graphs, and then determine the normality of a graph by computing node-level and graph-level reconstruction losses. However, these methods equally treat node-level anomaly and graph-level anomaly in the evaluation process of abnormal graphs, while the manifestation of anomalous graphs may vary in different graph sets. In other words, every type of graph data shows different degrees of node-level and graph-level anomalies. For instance, anomalous molecule graphs, differ from other normal molecule graphs in terms of the global structure, i.e., graph-level anomaly. But anomalous graphs are always node attribute anomalies or user connectivity anomalies in social network graphs, i.e., node-level anomaly. In addition, normal and abnormal graphs may be similar in some graph sets. If we only rely on the popular anomaly evaluation way based on graph representation reconstruction error, these abnormal graphs with subtle differences are difficult to be identified. Figure 1 illustrates a specific example. Thus, existing methods have not taken the above problems into account very well, and as a result, their performance in graph anomaly detection is also affected.

In this paper, a new approach for detecting the graph-level anomaly, namely \textbf{M}ulti-representations \textbf{S}pace \textbf{S}eparation based \textbf{G}raph-level \textbf{A}nomaly-aware \textbf{D}etection(MssGAD), is introduced to address the aforementioned challenges. In order to more effectively account for the varying significance of node-level and graph-level anomalies across different sets of graphs, we devise an anomaly-aware approach that dynamically adjusts the weighting of node-level and graph-level anomalies in the overall detection of graph anomalies. Furthermore, we learn a separate multi-representations space to differentiate normal graph representation space and abnormal graph representation space. Specifically, we select a certain number of graph samples from normal and anomalous graphs of the training set as anchor normal graphs and anchor abnormal graphs. Then, we train the proposed framework to make normal graphs closer to the anchor normal graph and anomalous graphs closer to the anchor anomalous graph under the weighted graph representation guidance. Thus, we can get contractible and separate anchor normal graph representation space and anchor abnormal graph representation space. Finally, we judge graph is normal or not by comparing the distances between it and the two kinds of graph representation spaces above. Therefore, the primary achievements of this research can be highlighted as:
\begin{itemize}
	\item [$\bullet$] We present a framework for detecting the graph-level anomaly through the use of the anomaly-aware strategy. For different graph sets, we dynamically adjust the weight of node-level anomaly and graph-level anomaly in the customized loss function optimization, to better adapt to different graph sets.
	\item [$\bullet$] We propose a multi-representation space separation method, which can gradually increase the distance between anchor normal graph representation space and anchor abnormal graph representation space during the training process. And we transfer the GLAD problem to the task of computing and comparing the distances between test graphs and two anchor graph representation spaces.
	\item [$\bullet$] We propose a new GLAD framework, namely MssGAD, which can effectively finish the task of graph-level anomaly detection. It also shows remarkable robustness to different datasets and it is not sensitive to the abnormal graph label in the collection.
\end{itemize}

The remaining sections of this paper are structured as follows: Section~\ref{sec2} provides an overview of related research. Section~\ref{sec3} elaborates on the problem definition of graph-level anomaly detection. Section~\ref{sec4} presents a comprehensive description of the implementation process and details of our framework. Section~\ref{sec5} showcases our method's experimental setup and results on various graph sets and compares our approach with some baseline methods. Finally, Section~\ref{sec6} provides concluding remarks.

\section{Related Work}\label{sec2}
With the continuous attention of researchers, graph neural networks have been developed significantly and many graph anomaly detection methods based on them have been proposed, and in this section, we present the related work from the above two aspects.
\subsection{Graph Neural Networks}
GNNs \cite{kipf2016semisupervised,velivckovic2017graph,wu2020comprehensive} are a type of neural network that can operate directly on graph-structured data. Conventional neural networks can handle vector or sequence information, and GNNs can learn and reason about relationships between objects represented as nodes in a graph and edges connecting them. GNNs have demonstrated significant promise in numerous tasks, including node classification \cite{gao2018large, rong2019dropedge, wang2020nodeaug}, link prediction \cite{zhang2018link, gou2022triad, chen2022gc}, and graph clustering \cite{zhang2018end, errica2019fair, lee2018graph}. A major benefit of GNNs is their capacity to grasp not only the local properties but also the global information of a graph, allowing them to make accurate predictions based on the structural characteristics of the graph.

There are various kinds of GNNs, including graph convolutional networks (GCNs) \cite{he2020lightgcn, wu2019simplifying, park2022acgcn,liao2022sociallgn}, graph attention networks (GATs) \cite{velivckovic2017graph,wang2019kgat}, and graph recurrent neural networks (GRNNs) \cite{hajiramezanali2019variational, geng2022graph}, each with its unique architecture and approach to graph learning. As interest in graph analysis continues to grow and graph-structured data becomes increasingly available, the use of GNNs is now crucial for data mining and many artificial intelligence topics will also utilize GNNs. We make use of GCNs in this paper to obtain node-level and graph-level representations of graphs, which enable us to identify graph anomalies from both local and global standpoints.

\subsection{Graph Anomaly Detection}
Graph anomaly detection involves identifying unusual patterns or behaviors in a dataset that is represented as a graph. This technique can be categorized into two types: graph-level anomaly detection (GLAD) and node-level anomaly detection (NLAD). Unlike traditional anomaly detection methods that operate on tabular or vector data, GLAD aims to identify the graphs which have unusual or unexpected patterns in a graph set. NLAD refers to identifying the nodes which are significantly different from others in a single graph.

Many GNNs-based graph anomaly detection approaches have been used and reached great performance. For instance, some NLAD methods \cite{tang2022rethinking, le2019probabilistic, jingcan2022gadmsl} concentrate on detecting abnormal nodes or edges within a vast graph. In addition, other NLAD methods \cite{chen2022gccad,jin2021anemone} consider building contrast instances to extract abnormal relationships of nodes. Recently, some GLAD methods such as  \cite{ma2022deep, luo2022deep, zhao2021using} focus on graph-level anomaly detection, and their performance is better than traditional shallow approaches \cite{liu2008isolation,breunig2000lof} on different graph sets. GLocalKD \cite{ma2022deep} is one example of a method that utilized graph representation distillation to obtain errors in node-level and graph-level representations, which are subsequently employed for the detection of anomalous graphs. OCGTL \cite{qiu2022raising} used advanced neural transformation learning to improve one-class classification performance in GLAD tasks. Other methods based on graph out-of-distribution detection are proposed, such as GOOD-D \cite{liu2023good} and  GraphDE \cite{li2022graphde}. 

But it should be noted that the manifestation of graph anomalies in different graph sets is not always consistent. Existing GLAD methods do not deal well with this problem, therefore, these methods failed to achieve good generalization performance. For example, the difference between anomalous and normal graphs in some datasets may specifically be more reflected at the node level, while in other datasets it is more reflected at the graph level. In response, we propose an anomaly-aware module that leverages dynamic factors to more accurately adjust the weight of graph-level and node-level anomalies in the overall detection of anomalies. And for the issue that anomalous patterns in some datasets are more subtle and difficult to grasp than others, we learn separate anchor normal graph representation space and anchor abnormal graph representation space to detect anomalous graphs more precisely. 

\section{Preliminaries}\label{sec3}
\subsection{Definition}
A graph set is represented as $\mathcal{G} = \{g_{1}, g_{2},..., g_{z}\}$, where z indicates the number of graphs in the dataset. And a single graph is denoted as $g = \{V, E, X\}$, where $V=\{v_{1}, v_{2},..., v_{n}\}$ denotes the set of nodes, $E = \{e_{1}, e_{2},..., e_{m}\}$ denotes the set of edges and $X = \{x_{1}, x_{2},..., x_{n}\}$ denotes the set of node attribute in the graph respectively. If the graph is attributed, the feature vector of node $v_{i}$, denoted by $x_{i}$, is generated by the node's attributes or labels. If the graph is not attributed, then $x_{i}$ represents the degree of node $v_{i}$. Moreover, $A$ is an adjacency matrix that denotes the topology of graph $g$, $A_{ij}$ is assigned 1 when the edge between $v_{i}$ and $v_{j}$ exists, and by contrast, 0 when not. 
\begin{itemize}
    \item \textbf{Node-level anomaly} means that a graph has unusual node attributes or labels and anomalous edges between nodes compared with nodes of normal graphs. We evaluate such anomalies by node-level representations. The node-level representations is the fusion feature distilled from node neighbors’ attributes and connection status between nodes which is denoted by $H_{V} = \{h_{v1}, h_{v2}, ..., h_{vn}\} \ vi\in V$.
    \item \textbf{Graph-level anomaly} indicates that there is an inconsistency in the overall structure of a particular graph compared to normal graphs. We capture this difference through graph-level representations. The graph-level representations concentrate on the more comprehensive information of the whole graph which is denoted by $H_{\mathcal{G}} = \{h_{g1}, h_{g2},...,h_{gz}\} \ gi\in \mathcal{G}$. In order to derive the graph-level representations, we apply a max-pooling operation to the node-level representations of each dimension for all nodes in the graph, and the specific formula is shown below:\begin{equation}
h_{g}=[\ \max_{i=1}^{n}\left(h_{vi}^{1}\right),\ \max_{i=1}^{n}\left(h_{vi}^{2}\right),\ ...,\ \max_{i=1}^{n}\left(h_{vi}^{d}\right)] \ \ vi\in g, \label{eq1}
\end{equation}
where the dimension of the node-level representations is d.
\end{itemize}

\begin{figure*}[!h]
	\small
	\includegraphics[width=\linewidth]{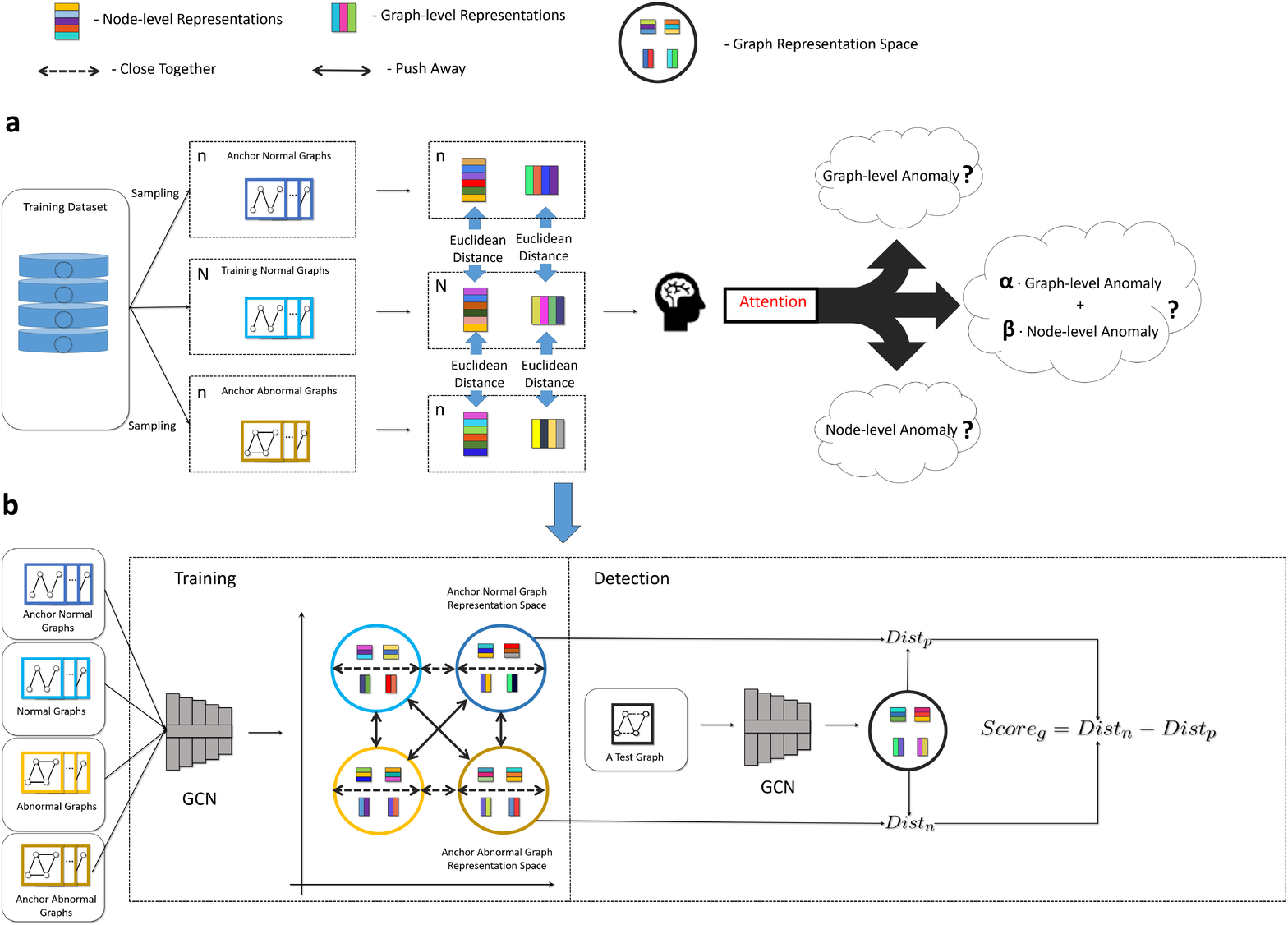}
	\caption{The framework of the proposed MssGAD. \textbf{a,} The Anomaly-aware Module. \textbf{b,} The Training and Detection Procedure. In the anomaly-aware module, we randomly select a certain ratio of anchor normal graphs and anchor abnormal graphs from normal graphs and abnormal graphs respectively. Then we calculate the node-level representations and graph-level representations of anchor normal graphs, anchor abnormal graphs, and training normal graphs. Next, we compute four Euclidean distances among the representations to determine the specific weight in terms of node-level and graph-level anomaly. In the training and detection procedure, we map the anchor normal graphs, the normal graphs, the abnormal graphs, and the anchor abnormal graphs to the multi-representations space. We promote the four graph representation spaces to close with or push away each other through the constraint of the loss function, and eventually make the normal and abnormal graph representation spaces separate from each other. During the reference process, the trained model is employed to map the test graph onto the graph representation space. Further, we measure the graph representation space distance between the test and anchor normal graphs as $Dist_{p}$ and obtain the graph representation space distance between the test and anchor abnormal graphs denoted by $Dist_{n}$ similarly. Eventually, we judge the label of the test graph by $Score_{g}$ which compares $Dist_{n}$ with $Dist_{p}$.}\label{fig2}
\end{figure*}
\subsection{Problem Statement}
Given a graph set $\mathcal{G} = \{g_{1}, g_{2},..., g_{z}\}$, where $g_{i} = \{V, E, X\}$ denotes an unlabeled graph that may be normal or abnormal, we aim to train a model which can proficiently identify anomalous graphs that demonstrate substantial deviations from the majority of normal graphs in both node-level and graph-level anomalies.  

\section{Methods}\label{sec4}
We introduce a framework for identifying graph-level anomalies in this section, which merges node-level and graph-level representations of graphs. The overall structure of the MssGAD framework we propose is illustrated in Figure 2. As a point of emphasis, we introduce the anomaly-aware module and the multi-representations space separation in detail and explain how they assist the model to identify more salient features. Then we describe the critical way to compute the score for anomaly detection.
\begin{table*}[!h]
\caption{Information and attributes of graph anomaly datasets.}
\resizebox{\linewidth}{!}{
\begin{tabular}{ccccccccccccc}
\hline
\textbf{Datasets} & \textbf{AIDS}   & \textbf{DHFR} & \textbf{MUTAG} & \textbf{PTC-FM} & \textbf{HSE} & \textbf{MMP} & \textbf{P53} & \textbf{PPAR} & \textbf{ENZYMES} & \textbf{PROTEINS}              \\ \hline
\textbf{GRAPHS} & 2000 & 756 & 188 & 349 & 8150 & 7320 & 8634 & 8184 & 600 & 1113 \\ 
\textbf{AVG-NODES} & 15.69 & 42.43 & 17.93 & 14.11 & 16.72 & 17.49 & 17.79 & 17.23 & 32.63 & 39.06 \\ 
\textbf{AVG-EDGES} & 16.2 & 44.54 & 19.79 & 14.48 & 17.04 & 17.83 & 18.19 & 17.55 & 62.14 & 72.82 \\ 
\hline
\end{tabular}
\label{tab1}
}
\end{table*}
\begin{table*}[!h]
\caption{The performance of anomaly detection in terms of mean AUC (\%) and standard deviation (\%). A = 0 means that graphs with label 0 in the dataset are treated as anomalous graphs, while A = 1 means that graphs with label 1 in the dataset are treated as anomalous graphs.}
\resizebox{\linewidth}{!}{
\begin{tabular}{ccccccccccccc}
\hline
\multirow{2}{*}{\textbf{Datasets}} & \multicolumn{2}{c}{FGSD-IF}               & \multicolumn{2}{c}{FGSD-LOF} & \multicolumn{2}{c}{FGSD-OCSVM} & \multicolumn{2}{c}{GOOD-D} & \multicolumn{2}{c}{OCGTL}       & \multicolumn{2}{c}{MssGAD}                \\
                                   & A=0                 & A=1                 & A=0           & A=1          & A=0            & A=1           & A=0           & A=1          & A=0                 & A=1        & A=0                 & A=1                 \\ \hline
\textbf{AIDS}                      & 99.35\tiny±0.75           & 0.65\tiny±0.75 & 87.92\tiny±3.45    & 12.08\tiny±3.45   & 83.47\tiny±4.89     & 16.50\tiny±4.89    & 92.62\tiny±2.19    & 14.56\tiny±7.39   & 94.89\tiny±3.74          & 98.90\tiny±1.13 & \textbf{99.61\tiny±0.76}          & 98.87\tiny±0.45          \\
\textbf{DHFR}                      & 51.51\tiny±4.09          & 48.49\tiny±4.09          & 49.05\tiny±5.97    & 50.95\tiny±5.97   & 55.05\tiny±6.37     & 44.95\tiny±6.37    & 62.82\tiny±2.98    & 61.35\tiny±4.73   & 71.77\tiny±8.78          & 50.15\tiny±6.22 & 72.24\tiny±2.05          & \textbf{72.27\tiny±2.87} \\
\textbf{MUTAG}                      & 78.98\tiny±8.67          & 21.02\tiny±8.67          & 74.45\tiny±5.94    & 25.55\tiny±5.94   & 57.70\tiny±5.71     & 42.30\tiny±5.71    & 79.41\tiny±2.45   & 18.64\tiny±8.66   & 86.41\tiny±9.34          & 82.05\tiny±9.56 & \textbf{90.07\tiny±3.42} & 89.18\tiny±4.74        \\  
\textbf{PTC-FM}                      & 40.86\tiny±5.48          & 59.14\tiny±5.48          & 42.88\tiny±4.83    & 57.12\tiny±4.83   & 50.38\tiny±7.88     & 49.62\tiny±7.88    & 54.85\tiny±5.76    & 51.37\tiny±5.45   & 60.52\tiny±5.06          & 46.56\tiny±3.97 & 61.58\tiny±3.59 & \textbf{62.19\tiny±5.00}          \\ \hline
\textbf{HSE}                       & 39.26\tiny±2.19 & 60.74\tiny±2.19          & 42.86\tiny±3.28    & 57.14\tiny±3.28   & 42.13\tiny±2.78     & 57.87\tiny±2.78    & \textbf{66.86\tiny±3.05}    & 55.98\tiny±3.37   & 64.98\tiny±4.34          & 64.57\tiny±4.06 & 55.76\tiny±7.26          & 52.87\tiny±6.85          \\
\textbf{MMP}                       & 67.97\tiny±1.08          & 32.03\tiny±1.08          & 57.71\tiny±1.78    & 42.29\tiny±1.78   & 51.90\tiny±2.04     & 48.10\tiny±2.04    & 70.14\tiny±2.09 & 52.85\tiny±3.44   & 61.01\tiny±1.34 & \textbf{80.53\tiny±1.27} & 68.07\tiny±0.75          & 68.05\tiny±0.81          \\
\textbf{P53}                       & 66.83\tiny±2.35          & 33.17\tiny±2.35          & 57.26\tiny±1.58    & 42.74\tiny±1.58   & 48.64\tiny±3.03     & 51.36\tiny±3.03    & 62.88\tiny±2.29          & 58.73\tiny±2.23   & 61.92\tiny±5.25          & \textbf{75.98\tiny±4.34} & 62.19\tiny±8.78          & 60.59\tiny±5.61 \\
\textbf{PPAR}                      & 34.83\tiny±3.10          & 65.17\tiny±3.10          & 47.14\tiny±4.96    & 52.86\tiny±4.96   & 50.13\tiny±5.27     & 49.87\tiny±5.27   & 66.61\tiny±1.55 & 55.01\tiny±4.07   & 61.07\tiny±4.10 & \textbf{70.36\tiny±6.58} & 61.04\tiny±7.04          & 65.76\tiny±3.94          \\ \hline
\textbf{ENZYMES}                   & 47.82\tiny±5.27          & 48.47\tiny±6.96          & 38.68\tiny±6.03    & 46.36\tiny±5.92   & 42.72\tiny±5.40     & 55.04\tiny±5.31    & 63.16\tiny±4.58    & 54.79\tiny±5.81   & 65.84\tiny±12.50          & 64.44\tiny±6.48 & \textbf{65.90\tiny±5.68} & 57.42\tiny±6.41          \\
\textbf{PROTEINS}                  & 75.30\tiny±2.21          & 24.68\tiny±2.21          & 58.29\tiny±3.74    & 41.71\tiny±3.74   & 31.81\tiny±0.83     & 68.19\tiny±0.83    & 72.20\tiny±4.00    & 72.32\tiny±3.32   & 59.34\tiny±1.27          & 64.78\tiny±2.96 & \textbf{77.24\tiny±3.04} & 76.63\tiny±3.41          \\ \hline
\end{tabular}
\label{tab1}
}
\end{table*}
\subsection{Anomaly-aware Module}
The key intuition of the anomaly-aware module is that the model is utilized to detect graph-level anomalies. The distinction between normal and abnormal graphs can be more specifically attributed to either the node level or the graph level. To account for this, we present an anomaly-aware module that assesses whether the anomalous pattern is mainly manifested in the node-level representations or the graph-level representations.

First, we divided the graph set $\mathcal{G}$ into two sets $G_{P}$ and $G_{N}$,$G_{P}=\{G_{P1},G_{P2},..., G_{Pn}\}\ G_{Pi}\in G$ contains normal graphs, and $G_{N}=\{G_{N1},G_{N2},...,G_{Nm}\} \ G_{Ni}\in G$ contains abnormal graphs. And we randomly extract a certain ratio of samples from $G_{P}$ and $G_{N}$ respectively denoted by $G_{PS}=\{G_{PS1},G_{PS2},...,G_{PSo}\} \ G_{PSi}\in G_{P}$ and $G_{NS}=\{G_{NS1},G_{NS2},...,G_{NSr}\} \ G_{NSi}\in G_{N}$. Under ideal conditions, the specific proportion of sampling should be determined by the distributions of graphs in the embedding space. However, the distributions are unknown. Therefore, after experiments with various parameters, we finally determine the ratio of sampling as follows:
\begin{equation}
\left|G_{s}\right|=4^{\lg\left(\left|G\right|\right)}\ \ \ {G_{s}\subset{G}},\label{eq2}
\end{equation}
where $\left|G_{s}\right|$ represents the number of anchor graphs, while $\left|G\right|$ corresponds to the total number of graphs within the set that have the same label. These samples are anchor graphs for the model during training. 

Subsequently, we utilize a graph neural network model to acquire both node-level and graph-level representations. We use the original graph-level and node-level representations of $G_{P}$, $G_{PS}$, and $G_{NS}$ to calculate four significant distances denoted by $D_{Pnode},\ D_{Pgraph},\\ D_{Nnode},\  D_{Ngraph}$:
\begin{equation}
D_{Pnode} = \frac{1}{\left|G_{P}\right|*\left|G_{PS}\right|}\sum\Vert f_{e}\left(h_{vi}\right)-f_{e}\left(h_{vj}\right)\Vert,\label{eq3}
\end{equation}
\begin{equation}
D_{Pgraph} = \frac{1}{\left|G_{P}\right|*\left|G_{PS}\right|}\sum\Vert H_{gi}-H_{gj}\Vert,\label{eq4}
\end{equation}
\begin{equation}
D_{Nnode} = \frac{1}{\left|G_{P}\right|*\left|G_{NS}\right|}\sum\Vert f_{e}\left(h_{vi}\right)-f_{e}\left(h_{vk}\right)\Vert,\label{eq5}
\end{equation}
\begin{equation}
D_{Ngraph} = \frac{1}{\left|G_{P}\right|*\left|G_{NS}\right|}\sum\Vert H_{gi}-H_{gk}\Vert,\label{eq6}
\end{equation}
\begin{equation}
{v_{i}\in G_{Pi},v_{j}\in G_{PSi},v_{k}\in G_{NSi},g_{i}\in G_{P},g_{j}\in G_{PS},g_{k}\in G_{NS}},\nonumber
\end{equation}
where $h_{vi}$ denotes the node representations of graphs in $G_{P}$, $h_{vj}$ denotes the node representations of graphs in $G_{PS}$ and $h_{vk}$ denotes the node representations of graphs in $G_{NS}$. Similarly, the graph-level representations of the graphs in the three aforementioned graph sets are respectively referred to as $H_{gi}$, $H_{gj}$, and $H_{gk}$. The function $f_{e}$ denotes the feature extraction process of node-level representations which can be common functions such as $Max\left(\right)$ and $Avg\left(\right)$, or more complex functions. In experiments, we use $Max\left(\right)$ as $f_{e}$. By combining both node-level representations and graph-level representations, the first two distances determine the dissimilarities between normal graphs and anchor normal graphs, while the latter two are relevant to normal graphs and anchor abnormal graphs. 

We refer to the Euclidean distance between $D_{Pnode}$ and $D_{Nnode}$ as the node-level representation difference and the Euclidean distance between $D_{Pgraph}$ and $D_{Ngraph}$ as the graph-level representation difference. These two distances assist us to evaluate which level of representation varies more significantly so that we can pay more attention to it during training. To quantify how much we focus on two levels of representations, we introduce two changeable weight factors $\alpha$ and $\beta$. Specifically, if the node-level representation difference is more than twice as large as the graph-level representation difference, $\alpha$ is empirically set to 0 while $\beta$ is set to 1, by which we emphasize the node-level representations when calculating the multi-representations space distances. On the contrary, if the graph-level representation difference is more than twice as large as the node-level representation difference, to utilize the graph-level representations, we set $\alpha$ to 1 and $\beta$ to 0. In other cases,  $\alpha$ and $\beta$ are equally set to 0.5 to fuse both of them. By using the anomaly-aware module to dynamically determine the weight factors, our model is able to adapt to different patterns of graph anomaly.

\subsection{Multi-representations Space Separation}
The motivation is the distribution of feature vectors in the embedding space can be very complex. The distribution of some anomalous graphs may be very close to that of normal graphs, making it difficult to discern subtle differences between them. Therefore the model may be misled and only achieve sub-optimal effects. To tackle this challenge, we utilize the extracted anchor graphs $G_{PS}$, and $G_{NS}$ in the anomaly-aware module to guide the normal and abnormal graphs to be separated.

We map the above four graph sets to the multi-representation space by fusing the graph-level representations and node-level representations in a weighted way. The specific graph representation space distances can be given as:
\begin{equation}
Dist_{1} = \alpha*\sum\Vert H_{gi}-H_{gj}\Vert+\beta*\sum\Vert f_{e}\left(h_{vi}\right)-f_{e}\left(h_{vj}\right)\Vert,\label{eq7}
\end{equation}
\begin{equation}
Dist_{2} = \alpha*\sum\Vert H_{gi}-H_{gk}\Vert+\beta*\sum\Vert f_{e}\left(h_{vi}\right)-f_{e}\left(h_{vk}\right)\Vert,\label{eq8}
\end{equation}
\begin{equation}
Dist_{3} = \alpha*\sum\Vert H_{gj}-H_{gk}\Vert+\beta*\sum\Vert f_{e}\left(h_{vj}\right)-f_{e}\left(h_{vk}\right)\Vert,\label{eq9}
\end{equation}
\begin{equation}
Dist_{4} = \alpha*\sum\Vert H_{gl}-H_{gj}\Vert+\beta*\sum\Vert f_{e}\left(h_{vl}\right)-f_{e}\left(h_{vj}\right)\Vert,\label{eq10}
\end{equation}
\begin{equation}
Dist_{5} = \alpha*\sum\Vert H_{gl}-H_{gk}\Vert+\beta*\sum\Vert f_{e}\left(h_{vl}\right)-f_{e}\left(h_{vk}\right)\Vert,\label{eq11}
\end{equation}
\begin{equation}
\begin{split}
&{v_{i}\in G_{Pi},v_{l}\in G_{Ni},v_{j}\in G_{PSi},v_{k}\in G_{NSi},}\\
&{g_{i}\in G_{P},g_{l}\in G_{N},g_{j}\in G_{PS},g_{k}\in G_{NS}}.\nonumber
\end{split}
\end{equation}
We can denote the graph representation space distance between normal graphs and anchor normal graphs by $Dist_{1}$, normal graphs and anchor abnormal graphs by $Dist_{2}$, anchor normal graphs and anchor abnormal graphs by $Dist_{3}$. Likewise, we can obtain the graph representation space distance between abnormal graphs and anchor normal graphs denoted by $Dist_{4}$ as well as which between abnormal graphs and anchor abnormal graphs denoted by $Dist_{5}$. 

During training, we can present the training loss function as:
\begin{equation}
Loss_{p}=Mean\left(Dist_{1}\right)-Mean\left(Dist_{2}\right)-Mean\left(Dist_{3}\right),\label{eq12}
\end{equation}
\begin{equation}
Loss_{n}=Mean\left(Dist_{5}\right)-Mean\left(Dist_{4}\right)-Mean\left(Dist_{3}\right).\label{eq13}
\end{equation}
In each epoch, we train the graph neural network model with normal graphs by minimizing the $Loss_{p}$, and then with abnormal graphs by the $Loss_{n}$. So that the model can learn every graph in the training set. And we drive the anchor normal graphs and the anchor abnormal graphs apart by the constraint of $Mean\left(Dist_{3}\right)$ in the loss function and then make normal graphs close to anchor normal graphs under the action of $Mean\left(Dist_{1}\right)$ while far away from anchor abnormal graphs due to $Mean\left(Dist_{2}\right)$. Symmetrically, the abnormal graphs will approach the anchor abnormal graphs because of $Mean\left(Dist_{5}\right)$ and escape from the anchor normal graphs as a result of $Mean\left(Dist_{4}\right)$. Generally, the multi-representations space distance between normal graphs and abnormal graphs will vary to be bigger. So it is less difficult for the model to learn their feature distinctions. And during inference, the performance of the anomaly score based on multi-representations space distances will be better as well.

\subsection{Graph Anomaly Evaluation}
The graph anomaly evaluation is mainly based on the multi-represen-tations space distance between the unlabeled graph set and anchor normal graphs as well as the one between the unlabeled graph set and anchor abnormal graphs. We don’t take the whole normal graphs set or the abnormal graphs set into account. Because our anomaly score is dependent on the distance, we hope to select the graphs with more typical features, namely the anchor graphs, to reduce the impact of the singular graphs and improve the precision. 

Specifically, to detect abnormal graphs with the trained model, we can calculate two vital graph representation space distances of a given graph $G_{un}$. $Dist_{p}$ denotes the graph representation space distance between $G_{un}$ and $G_{PS}$, and $Dist_{n}$ denotes the graph representation space distance between $G_{un}$ and $G_{NS}$ as follows:
\begin{equation}
\begin{split}
Dist_{p} = \alpha*\sum\Vert H_{gi}-H_{gj}\Vert+\beta*\sum\Vert f_{e}\left(v_{i}\right)-f_{e}\left(h_{vj}\right)\Vert\\
{v_{i}\in G_{un},v_{j}\in G_{PSi}},\label{eq14}
\end{split}
\end{equation}
\begin{equation}
\begin{split}
Dist_{n} = \alpha*\sum\Vert H_{gi}-H_{gk}\Vert+\beta*\sum\Vert f_{e}\left(h_{vi}\right)-f_{e}\left(h_{vk}\right)\Vert\\{v_{i}\in G_{un},v_{k}\in G_{NSi}}.\label{eq15}
\end{split}
\end{equation}
\begin{equation}
Score_{g}=Dist_{n}-Dist_{p}.\label{eq16}
\end{equation}
If $G_{un}$ belongs to the normal graphs set, $Score_{g}$ will be greater than a certain threshold value, on the contrary, $Score_{g}$ being smaller than the threshold value indicates $G_{un}$ is more likely to be anomalous. Empirically and ideally, we use 0 as the threshold value. Finally, the algorithm \ref{algorithm1} shows the detailed training process of MssGAD.

\begin{algorithm}[h!]
\caption{MssGAD}
\label{algorithm1}
\KwIn{graph set $\mathcal{G}$; normal graph set $G_{P}$  and abnormal graph set $G_{N}$, $\mathcal{G}=G_{P} \cup G_{N}$; GCN  model $M$; unlabeled graph sample $G_{un}$; training epochs $E$ and training batches $B$;}
\KwOut{anomaly score $\mathbf{Score_{g}}$.}
Randomly initialized model weights of $M$; \\
Calculate the count of anchor normal graphs as $S_{P}$ and the count of anchor abnormal graphs as $S_{N}$ by Eq.\eqref{eq2};\\
Randomly draw an anchor normal graph set $G_{PS}$ with a size of $S_{P}$ from $G_{P}$ and an anchor abnormal graph set $G_{NS}$ with a size of $S_{N}$ from $G_{N}$;\\
Calculate $D_{Pnode}$, $D_{Pgraph}$, $D_{Nnode}$, $D_{Ngraph}$ by Eq.\eqref{eq3}, Eq.\eqref{eq4}, Eq.\eqref{eq5}, Eq.\eqref{eq6};\\
Determine the value of $\alpha$ and $\beta$ by $D_{Pnode}$, $D_{Pgraph}$, $D_{Nnode}$, $D_{Ngraph}$;\\
Use graph set $\mathcal{G}$ as training set;

\For{$i\gets1$ \KwTo $E$}{
    Use normal graph set $G_{P}$ as training set;\\
    \For{$j\gets1$ \KwTo $B$}{
    Distill $H_{v}$ and $H_{g}$ from $M$ by Eq.\eqref{eq1}; \\
    Calculate $Dist_{1}, Dist_{2}, Dist_{3}$ on the normal graph set $G_{P}$ by Eq.\eqref{eq7}, Eq.\eqref{eq8}, Eq.\eqref{eq9}; \\
    Update $M$ by minimizing $Loss_{p}$ shown in Eq.\eqref{eq12};
    }
    Use abnormal graph set $G_{N}$ as training set;\\
    \For{$j\gets1$ \KwTo v$B$}{
    Distill $H_{v}$ and $H_{g}$ from $M$ by Eq.\eqref{eq1}; \\
    Calculate $Dist_{3}, Dist_{4}, Dist_{5}$ on the normal graph set $G_{N}$ by Eq.\eqref{eq9}, Eq.\eqref{eq10}, Eq.\eqref{eq11}; \\
    Update $M$ by minimizing $Loss_{n}$ shown in Eq.\eqref{eq13};
    }}
Use unlabeled graph sample $G_{un}$; \\
Distill $H_{v}$ and $H_{g}$ from $M$ by Eq.\eqref{eq1}; \\
Calculate $Dist_{p}, Dist_{n}$ on the unlabeled graph sample $G_{un}$ by Eq.\eqref{eq14}, Eq.\eqref{eq15}; \\
Calculate $Score_{g}$ on unlabeled graph sample $G_{un}$ by Eq.\eqref{eq16}; \\ 

\textbf{Return} anomaly score $\mathbf{Score_{g}}$.
\end{algorithm}

\section{Experiment}\label{sec5}
In this section, we present an evaluation of the MssGAD model on diverse publicly available datasets from various domains and compare its performance with several other anomaly detection methods proposed in recent years. Based on this, we then design and conduct several comparative experiments and try to interpret and validate the effectiveness of our model design through the results of these experiments.
\subsection{Datasets}
For the selection of datasets, ten different datasets are selected from multiple fields such as toxic substances, molecular compounds, macromolecular proteins, etc. HSE, MMP, p53, and PPAR are datasets established to predict the properties and effects of compounds according to their chemical structures, which contain thousands of potentially harmful chemicals. AIDS is an antiviral screening dataset used for NCI / NIH development and treatment programs. DHFR is a dihydrofolate reductase inhibitors dataset that contains structural information for 325 compounds. For each compound, 228 molecular descriptors have been calculated. PTC-FM contains compounds that are carcinogenic markers for rodents. The ENZYMES and PROTEINS datasets are composed of protein structures, where graphs correspond to individual proteins and nodes correspond to secondary structure elements. Two nodes are connected by an edge if they are adjacent in either the amino acid sequence or 3D space. Table 1 presents the specifics of these datasets. These datasets are not identical in structure because they belong to different domains, some of them contain node features and others do not, therefore we use the degrees of the nodes to replace node features when they do not exist. In addition, we will take one class of graphs as anomalous graphs, while the graphs of the other categories will be treated as normal graphs during experiments.                             
\subsection{Baselines}
For the choice of comparing methods, we select three more traditional methods: FGSD-IF \cite{liu2008isolation}, FGSD-LOF \cite{breunig2000lof}, and FGSD-OCSVM \cite{scholkopf1999support}, which are commonly based on the Family of Graph Spectral Distances (FGSD) \cite{verma2017hunt}. FGSD is a method to embed node pairwise distances and obtain the graph feature representations which exhibits certain stability and uniqueness, while isolation forest (IF), local outlier factor (LOF), and one-class support vector machine  (OCSVM) are three classical classifiers combined with FGSD. Furthermore, GOOD-D \cite{liu2023good} is selected because it is a contrastive learning-based unsupervised graph anomaly detection approach that has been recently introduced. In addition, OCGTL \cite{qiu2022raising} is a graph-level anomaly detection method that incorporates concepts from transformation learning and self-supervised learning, and has recently set a new standard in this area. To make MssGAD more convincing, we also added OCGTL to the baseline methods for comparison.
\subsection{Experimental Implementation}
To measure the effectiveness of our method more comprehensively and accurately, we test each dataset twice, regarding the graphs with label 0 and the graphs with label 1 as anomalous graphs respectively while the rest graphs of the dataset as normal graphs, and employ 5-fold cross-validation in each experiment to obtain more reliable and consistent outcomes. In our experiments, we assess the performance based on the mean AUC and its standard deviation, as the evaluation criterion. It is also worth mentioning that GOOD-D provides detailed configurations of training hyperparameters on all datasets except for the MUTAG and PTC-FM datasets, so we determine the training hyperparameters on these two datasets by ourselves.
\subsection{Anomaly Detection Performance}
The specific anomaly detection results of each baseline on the above ten datasets are shown in Table 2. A=0 and A=1 represent the cases when the label of anomaly graphs is set to 0 and 1 respectively. The data presented in the table reveals that the MssGAD model performs better than most of the other methods on the majority of the datasets, specifically, on the data sets of AIDS, DHFR, MUTAG, PTC-FM, ENZYMES, and PROTEINS than several of the selected baseline methods. While on the remaining four datasets of HSE, MMP, P53, and PPAR, MssGAD performs less well than GOOD-D and OCGTL. Among them, DHFR, ENZYMES, and PROTEINS have more average number of nodes and edges, but our model still performs the detection task relatively well, which we believe is due to our multi-representations space separation method. It allows us to still discriminate the difference between normal graphs and abnormal graphs clearly in a relatively high-dimensional representation space. As for the four datasets with poorer results, the reason is that as the size of the dataset increases the number of anchor graphs also becomes larger, resulting in the distance-based loss function becoming less effective in the case of more complex distributions. The randomly selected graphs do not perform well enough as anchor graphs. Additionally, by comparing the values, it can be found that the accuracies of these two methods usually differ greatly between the two cases of A=0 and A=1. And there is often a difference of more than ten percentage points between the lower accuracy and the higher accuracy. The reason is that the graph set usually consists of more than two varieties of graphs in many cases. We commonly regard one class as the abnormal graph set and the others compose the normal graph set. Due to the diversity of categories in the normal graph set, the distribution of the normal graphs in the embedding space tends to be more scattered than the abnormal graphs. The haphazard distribution limits the performance of some existing methods when they try to distill consistent feature representations of the normal graphs. Thus their performances vary widely and can not be very stable across data sets and anomalous labels. Thanks to our symmetrical sampling from both normal and abnormal graph sets, the distributions of both the normal graph set and abnormal graph set become tighter with the bootstrap of the anchor graphs and the constraint of the loss function. MssGAD can better grasp the pattern of both sets so as to maintain relatively close accuracies no matter A=0 or A=1. Therefore, MssGAD can achieve relatively good anomaly detection results on most of the datasets.

\setlength{\parskip}{-1pt}

\subsection{Ablation Experiments}
In addition to comparing the performance with the baseline approaches, we also conduct ablation experiments. And then analyze the results with the key design points in our work.

The first part is the ablation experiment for the anomaly-aware module. As mentioned in the previous section, to take into account both graph-level representations and node-level representations comprehensively, we introduce two dynamic factors $\alpha$ and $\beta$. Their values will be determined by comparing the differences between the two kinds of representations, in order to emphasize the representations with more obvious distinction. In the ablation experiment, we set both $\alpha$ and $\beta$ to 1, then divide the experiment into two groups with the anomalous label being 0 and 1. And test on the previously selected datasets and analyze the results with those in the previous baseline experiment. The results of our experiments are displayed in Figure 3. The solid bar presents the mean value of AUC while the top line reflects the standard deviation of AUC. The red one is the original model and the green one is the ablation model. In general, the results suggest that the original model achieves better performance than the ablation model. The average AUC of the ablation model appears to be lower than the original model across datasets. Precisely, the differences are larger in some datasets and smaller in others. We consider that the reason is that setting $\alpha$ and $\beta$ to constant values may happen to reflect the anomalous patterns of the graph reasonably, however, they don't fit on other datasets. It indicates that the anomaly-aware module does help the model to sense and refine more representative anomalous pattern handling various datasets.
\begin{figure}[h!]
\includegraphics[width=1.0\linewidth]{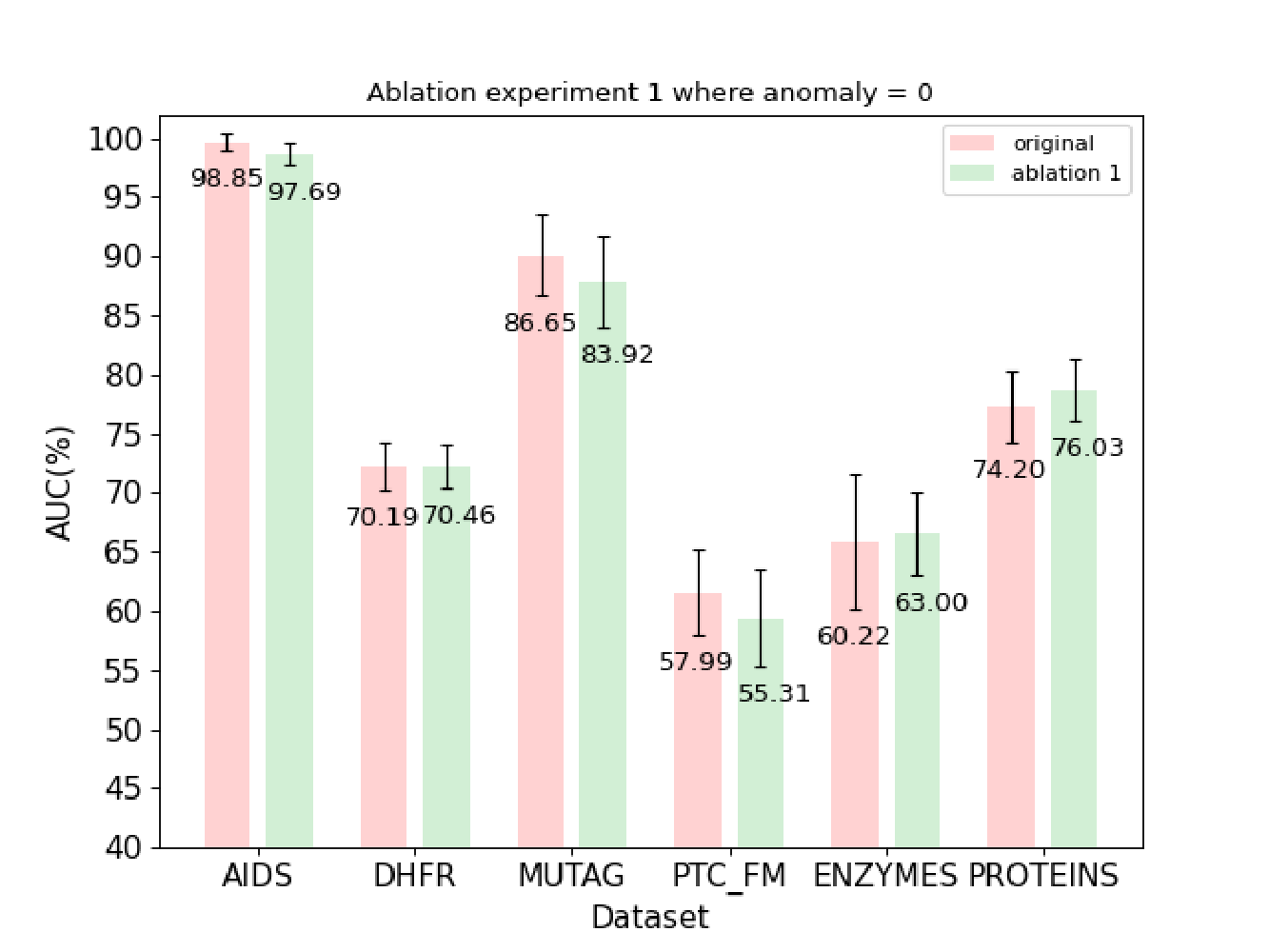}  
\includegraphics[width=1.0\linewidth]{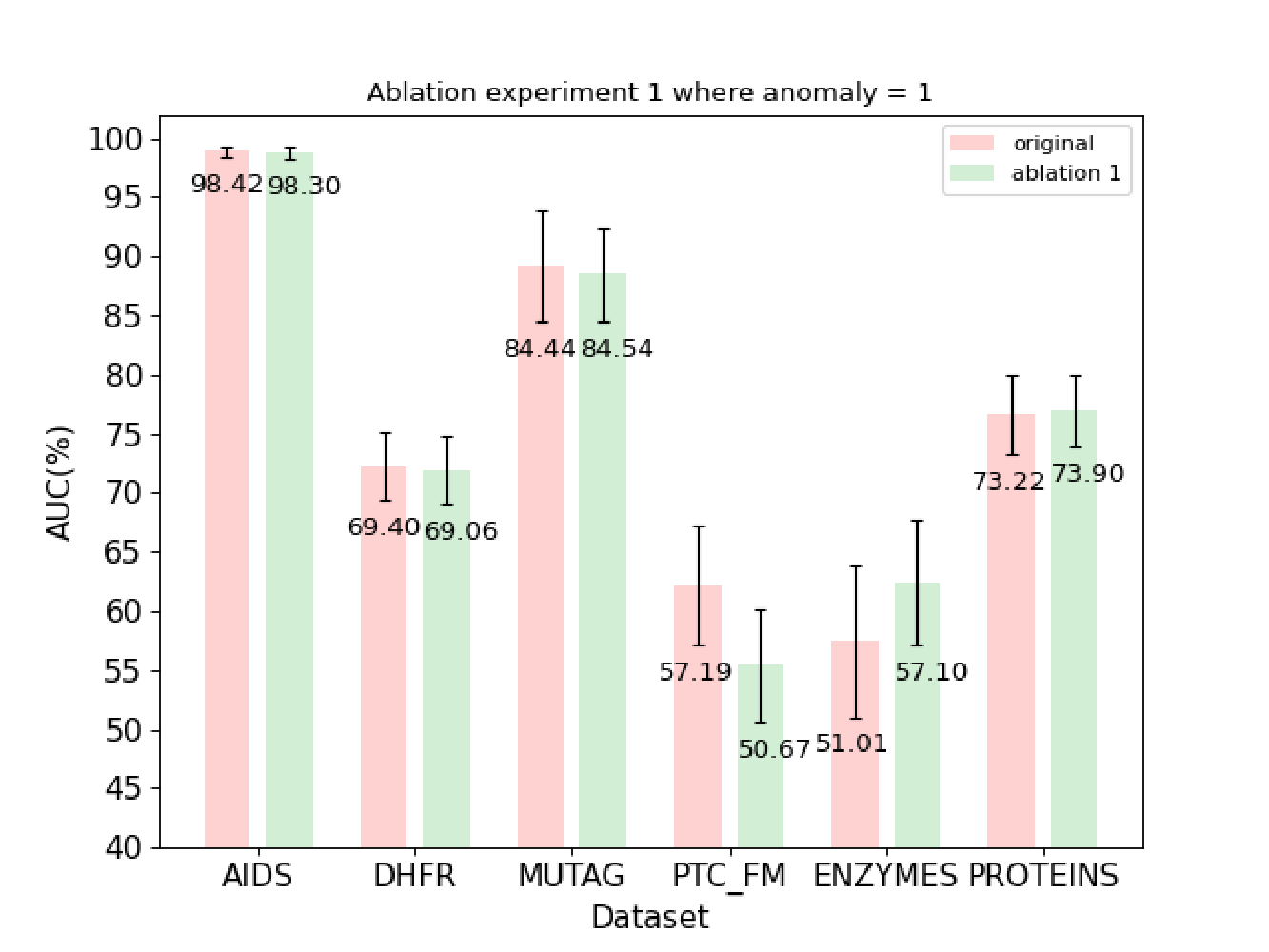}
\caption{The AUC results of ablation experiment for the anomaly-aware module. The bars represent the AUC values of each model, and the top lines show the standard deviation. Anomaly = 0 means the graphs whose label is 0 are regarded as anomalous graphs. Similarly, anomaly = 1 means the graphs whose label is 1 are regarded as anomalous graphs.}    
\label{fig3}       
\end{figure}
\begin{figure}[h!]
\includegraphics[width=1.0\linewidth]{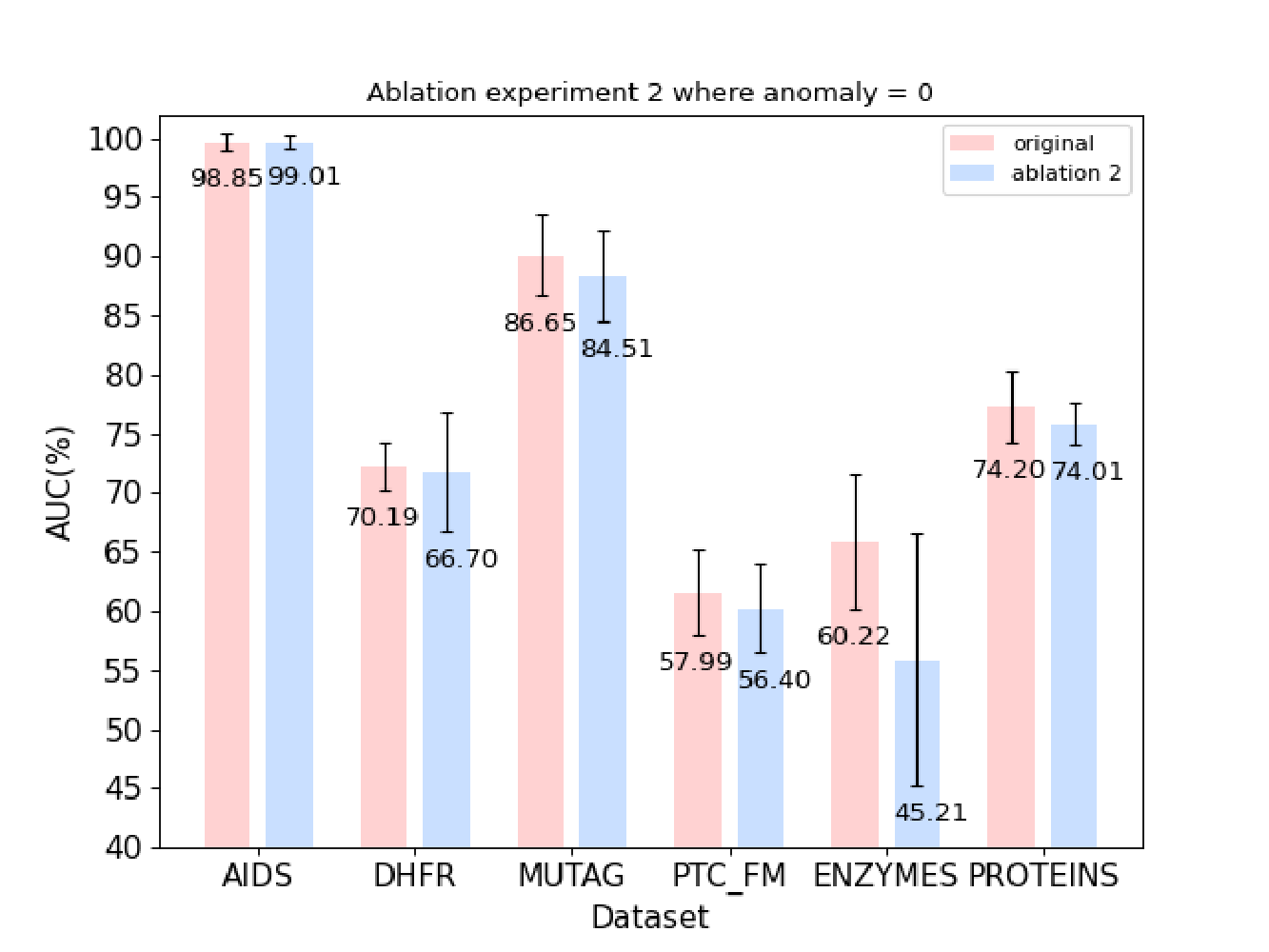}
\includegraphics[width=1.0\linewidth]{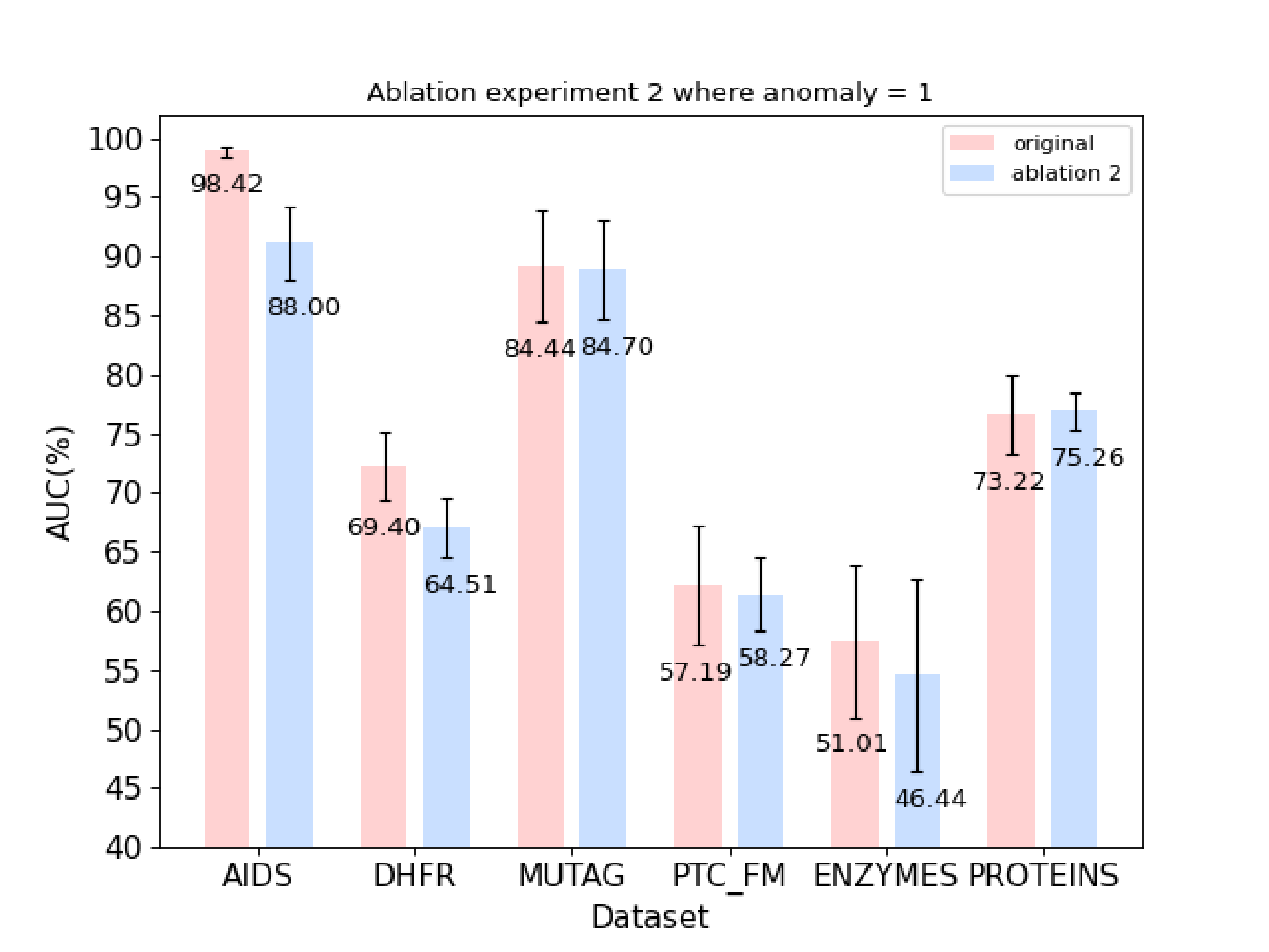}
\caption{The AUC results of ablation experiment for anchor graphs. The bars represent the AUC values of each model, and the top lines show the standard deviation. Anomaly = 0 means the graphs whose label is 0 are regarded as anomalous graphs. Similarly, anomaly = 1 means the graphs whose label is 1 are regarded as anomalous graphs.}   
\label{fig4}       
\end{figure}

The second part is the ablation experiment designed for investigating the role played by the anchor graphs. We randomly select anchor graphs from the normal and abnormal graph sets respectively. And we hope that the unselected graphs of the same label gradually converge in the multi-representations space with the anchor graphs as the center during training. And we guide the anchor normal graphs and anchor abnormal graphs to gradually separate from each other through the restriction of $Mean\left(Dist_{3}\right)$. Eventually, the normal and abnormal graphs are separated along with the anchor normal graphs and anchor abnormal graphs. Therefore, we remove $Mean\left(Dist_{3}\right)$ from $Loss_P$ and $Loss_N$ in the second ablation experiment, maintaining only the rest parts, and then experiment again. Figure 4 displays the comparison results. The red one is still the original model, while the blue one is the ablation model. In comparison to the original model, the ablation model exhibits reduced accuracy. And there is a large increase in the standard deviation of the AUC of the ablation model on some datasets. This indicates that the performance of the model in extracting anomalous patterns becomes more unstable. We believe this is because the unseparated normal graph representations and abnormal graph representations are more difficult to find the boundaries of the division. And the comparison clearly shows that $Mean\left(Dist_{3}\right)$ helps to expand the graph representation space distance between anchor normal graphs and anchor abnormal graphs. The distinction can be more easily learned by the model after being expanded. Thus the original model can refine the subtle feature differences and obtain a better performance, which fulfills our idea of design.

In addition, we can see that the variances of the results for MUTAG, PTC-FM, and ENZYMES are relatively large, considering that the reason is the small number of graphs in the datasets which lead to the fluctuations in the quality of the extracted anchor graphs. Similarly, the greater variances of the original model results on some datasets are due to unreasonable weight values determined by the anomaly-aware module based on unreliably anchor graphs in some cases, but the overall average results are still better.

\begin{figure}[!h]
\subfigure[AIDS/A=0]{  
\includegraphics[width=4cm]{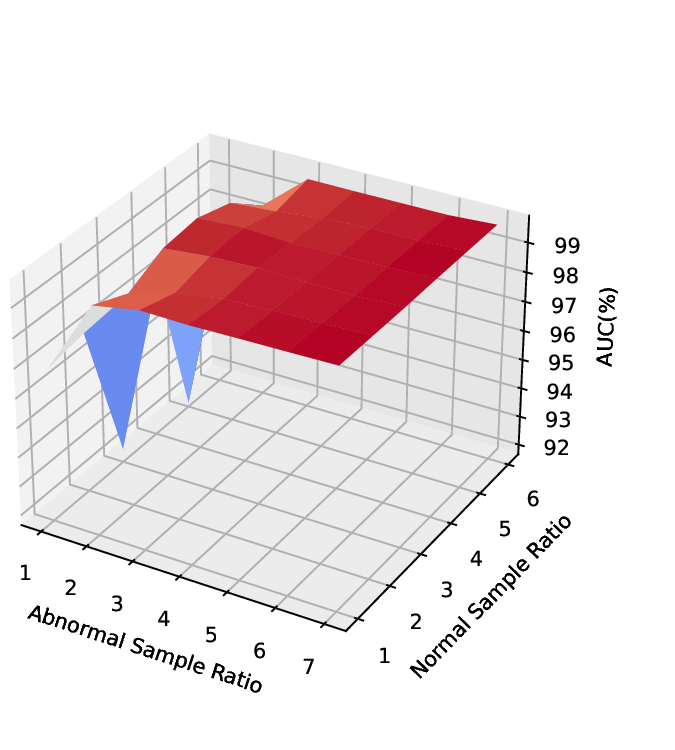}  
}
\subfigure[AIDS/A=1]{ 
\includegraphics[width=4cm]{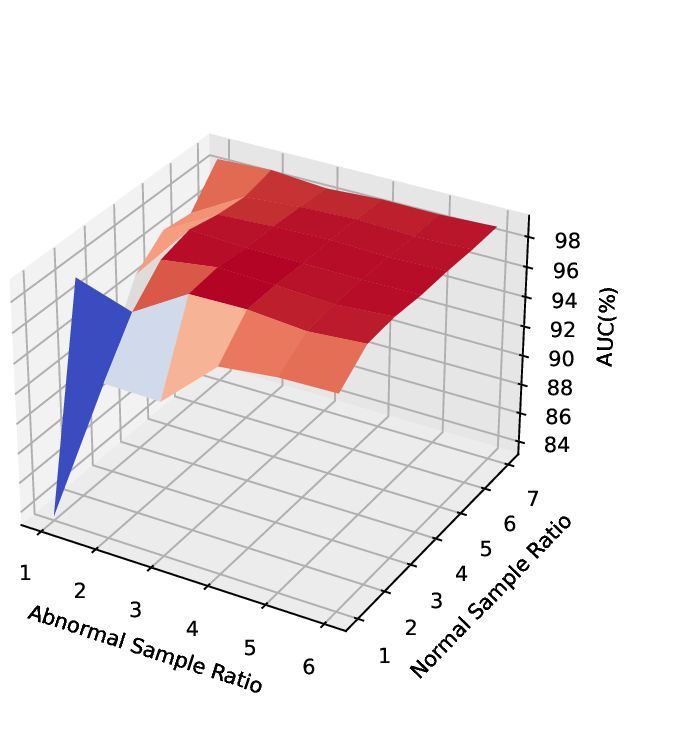}
}

\subfigure[DHFR/A=0]{  
\includegraphics[width=4cm]{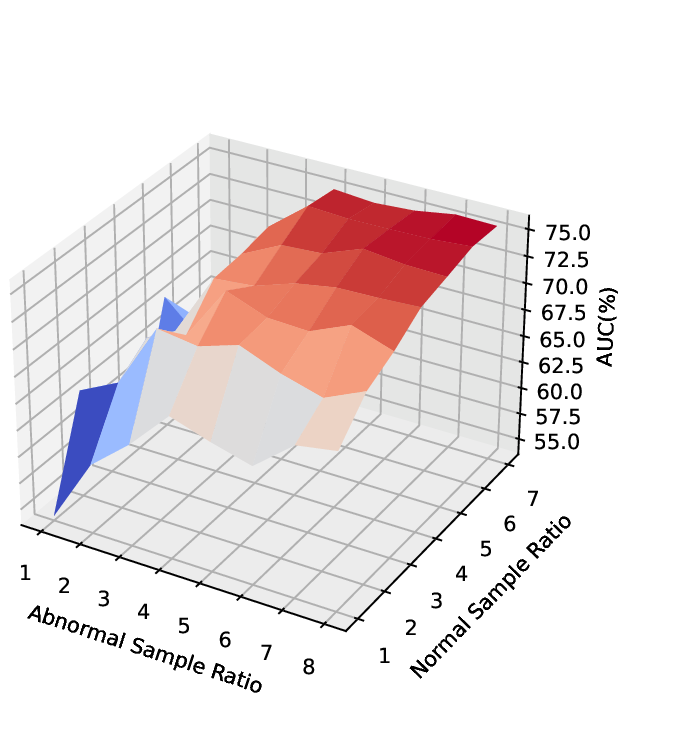}
}
\subfigure[DHFR/A=1]{ 
\includegraphics[width=4cm]{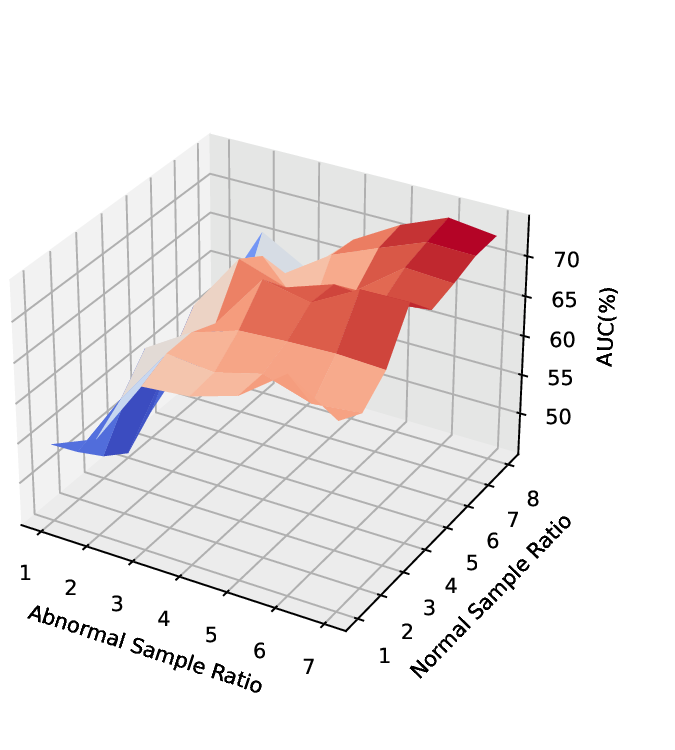}
}

\subfigure[MUTAG/A=0]{  
\includegraphics[width=4cm]{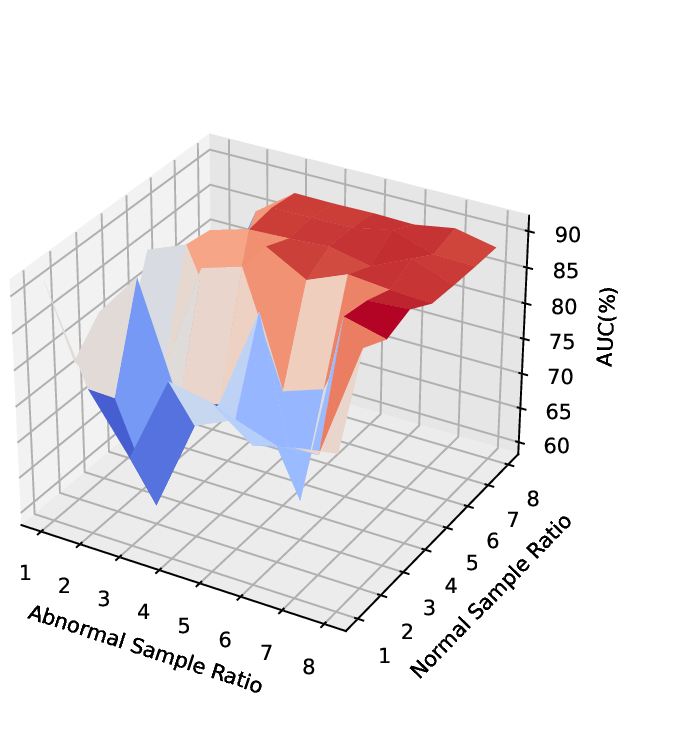}
}
\subfigure[MUTAG/A=1]{  
\includegraphics[width=4cm]{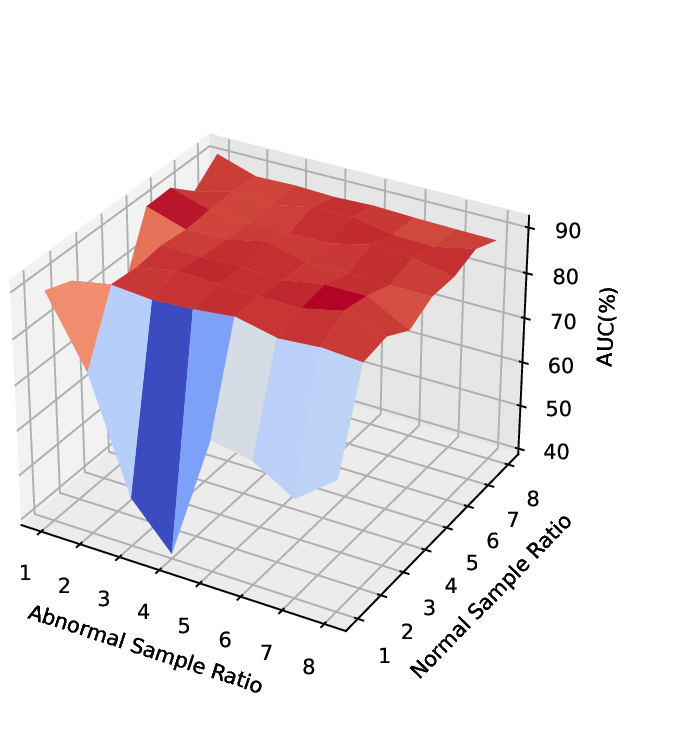}
}
\caption{The AUC results of MssGAD under different ratios of samples in the normal graph set and abnormal graph set. A = 0 denotes that the graphs with label 0 in the dataset are anomalous. Similarly, A = 1 denotes that the graphs with label 1 in the dataset are anomalous.}    
\label{fig5}       
\end{figure}

\subsection{Parameter Analysis}
In the previous ablation experiments, we initially illustrate the effectiveness of the anchor graphs. So in this session, we experiment and analyze one of the crucial parameters, i.e., the sampling ratio. In this session, we experiment with different normal and abnormal sampling ratios. In detail, we take the sampling ratio factor k as 1, 2, 3, 4... and so on consecutive integers up to the number of the whole graph set. And the relationship between k and the anchor graph set $G_{s}$ can be given as follows:
\begin{equation}
\left|G_{s}\right|=k^{\lg\left(\left|G\right|\right)}\ \ \ {G_{s}\subset{G}},\label{eq17}
\end{equation}
where G refers to the set of all normal graphs or the set of all abnormal graphs. Since there is a rounding operation when taking logarithms, the maximum value that k can reach needs to be determined according to the number of graphs in the whole set. During the experiments, we set various sampling ratio factors on the normal and abnormal graph set and then use 5-fold cross-validation. Our experimental evaluation involves three datasets, namely AIDS, DHFR, and MUTAG, for a total of six scenarios with anomalous labels of 0 and 1, respectively. Figure 5 illustrates that the trends of AUC vary across different datasets and abnormal labels. However, as the ratio of normal and abnormal sampling ratio is relatively small to relatively large, the overall AUC shows an upward trend. But when the ratio reaches a certain level, there is no obvious improvement continuing increasing the ratio. Therefore, we analyze that when the random sample is too small, the selected anchor graphs can not represent the features of the overall dataset properly; however, if the sampling ratio is too high, it will lead to a great increase in computation, so we finally set the final sampling ratio factor to 4 to achieve a balance between the two aspects.

\section{Conclusion}\label{sec6}
In this work, we propose MssGAD to solve two common problems encountered in graph-level anomaly detection. We present the anomaly-aware module as a solution to the problem that anomalous graphs in different datasets tend to exhibit varying emphasis on node-level and graph-level anomalies. We dynamically adjust the weight factors according to the magnitude of the morphological differences between node-level and graph-level representations, enabling a more accurate and flexible fusion of both two representations. Further, to deal with the problem that it is hard to distinguish anomalous graphs from normal graphs when they are very similar, we expand the gap between the normal and abnormal graphs based on multi-representations space separation. So that we can precisely detect every anomalous graph by its distance error with anchor graph representation spaces. The experiments show that MssGAD performs well on most of the publicly available datasets, and the subsequent ablation and comparison experiments also demonstrate that the design of the model achieves the expected results.

\begin{acks}
We would like to thank the anonymous reviewers for their helpful comments and feedback. This research was supported by Wuhan University People's Hospital Cross-Innovation Talent Project Foundation under JCRCZN-2022-008. 
\end{acks}


\end{document}